\documentclass[a4paper]{article}
\usepackage[margin=0.75in]{geometry}

\usepackage{amssymb}
\usepackage{graphicx}

\usepackage{url}
\urldef{\mailtb}\path|thomas.burger@cea.fr|

\title{Bridging belief function theory to modern machine learning}
\author{Thomas Burger\footnote{The author would like to thank Fabio Cuzzolin, who organized, a ``\emph{discussion panel on the status and future of the belief function theory}'' during the BELIEF 2014 conference~\cite{belief2014}. This very enriching brainstorm was the starting point for this work, which can be seen as a rejoinder to this discussion panel.}\\
iRTSV-BGE (Universit\'e Grenoble-Alpes, CNRS, CEA, INSERM)\\ Grenoble, France\\ \mailtb
}

\begin{document}
\maketitle


\abstract{
Machine learning is a quickly evolving field which now looks really different from what it was 15 years ago, when classification and clustering were major issues. This document proposes several trends to explore the new questions of modern machine learning, with the strong afterthought that the belief function framework has a major role to play.}\\

\noindent \textbf{Keywords:} belief functions; machine learning;

\section{Introduction}
In an age of user generated web-contents and of portable devices with embedded computer vision capabilities, machine learning (ML) and big data mining questions are fundamental. As a result, these questions naturally penetrate neighboring research fields, including belief function theory (BFT), so that it is now usual to attend a ``Classification'' session~\cite{belief2012} or a ``Machine Learning'' session~\cite{belief2014} in a conference devoted to belief functions. 

However, it is hard to accept that among the various proposed approaches based on BF, very few have become state-of-the-art ML methods, the knowledge of which has spread beyond the BF community. Without any doubt, this can be partly explained by the relative size of the scientific communities under consideration: although quickly growing, the BF one is relatively small with respect to that of statistics, Bayesian networks, neural networks, etc. 
However, this reason alone is not sufficient: There are indeed other topics, such as for instance, information fusion, where BF-based methods are now as well recognized as are methods based on more classical formalisms, such as probabilities, or ontologies.

In this report, I assume an additional reason: that some researchers focused on BFT (especially the youngest), who have progressively turned their interests towards ML problems, may not capture the newest trends of this field. In fact, I used to be an example of such researchers, and I acknowledge that my first perceptions of ML were clearly outdated. This is why, I propose a short review of the respective evolution of BFT and of ML, as well as an attempt to put them in perspective. Of course, many senior researchers may find this exercise futile, as they have their own broad view on the question. However, to my knowledge, no recent referenced article is available for any reader seeking for a starting point to question the links between ML and BFT.

This document is structured as follow: In Section~\ref{sec:landscape}, a brief recall of the evolution of the mainstream in the BF community is provided. Then, in Section~\ref{sec:class-clus}, a short summary of the earlier ages of ML up to the mid-90s, is sketched, as well as a coarse description of the successful interactions between ML and BFT in those times. Afterward, I provide in Section~\ref{sec:mml} a synthetic overview of the revolution that blew over ML around the early 2000s, and which modified its goals and the organization of its supporting community. As BFT does not seem to fit in this new picture of the ML world, I list in Section~\ref{sec:room} a few problems that may still be of interest for the current mainstream of BFT, as well as some potential interesting evolutions for the community to adapt to the newly raised questions.

\section{The landscape of belief function interpretations}\label{sec:landscape}
When considering BF interpretations, one often opposes Demspter's \emph{imprecise statistic view} to Shafer and Smets \emph{singular view}, such as described in~\cite{destercke2013}. However, there is another way to sort the various interpretations: it is to refer to the mathematical object a mass function is perceived as a generalization of. This leads to the following taxonomy:
\begin{itemize}
	\item The \textbf{probability-affiliated} interpretation: ``A mass function is an object which generalizes a discrete probability measure, where the probability masses are not necessarily known, but are only assumed to belong to an interval''. Whatever the origin of this imprecise probability, i.e. either statistical (such as in Dempster's view~\cite{Dempster67,dempster1968generalization}, or as in the Theory of Hints~\cite{kohlas1995mathematical}), or subjective (such as with Shafer or Shenoy~\cite{Shafer76,shenoy1986propagating,shenoy1990axioms,shenoy1992valuation}), the mathematical theory behind is that of random sets~\cite{kendall1974foundations,nguyen2006introduction}.
	\item The \textbf{set-affiliated} interpretation: ``A mass function is a set description of the real value of an ill-known variable, which is enriched by weightings which add up to one''. Of course, this view is conceptually closer to that of fuzzy sets, possibility theory~\cite{dubois1998possibility}, and to artificial intelligence (AI) in general. 
	\item The \textbf{capacity-affiliated} interpretation: ``A mass function is only a particular type of Choquet's capacity, the Moebius transform of which is totally monotone''. Such interpretation is strongly related to multi-criteria optimization and to operational research~\cite{grabisch1996application}.
\end{itemize}

From a historical perspective, the oldest probability-affiliated interpretation (of Dempster) was rooted in imprecise statistics, before moving towards non-frequentist views (that are more classicaly associated to nowadays subjective views), under the influence of Shafer; yet, in Shafer and Shenoy works~\cite{shafer1991local}, the link to probability remains strong as the reference to random set is explicit. Later on, the set-affiliated interpretation has spread under the influence of renown authors (such as Zadeh, Dubois, Smets or Yager), with a strong background in data fusion, fuzzy sets and expert systems. Among them, Smets constantly positioned his work with respect to Bayesian classifiers~\cite{ristic2004kalman} or Bayesian reasoning~\cite{smets2005decision}; however, the Transferable Belief Model (TBM~\cite{smets94a}) was a non-probabilistic model, which, in Smet's view~\cite{smets1992TBMrandomset}, was not compliant with random sets. 
Besides, the capacity-affiliated interpretation of BF has never really developed on its own, independently of other works in operational research and in game theory. Finally, todays, the great majority of the belief function community accepted the set-affiliated interpretation, as it appears in proceedings such as~\cite{belief2012,belief2014}, where the majority of the  articles undertook this interpretation.

\section{Classification, clustering and belief functions}\label{sec:class-clus}
As fully described by Miclet and Cornu\'ejols in~\cite{cornuejols2009place}, the short history of ML is full of rapid and strong evolutions, so that to date, the discipline which was originally part of AI, has completely separated from its parent discipline to make its own path, which appears to strongly converge toward statistics and optimization: They review the original heuristics motivated by bio-inspiration (such as multi-layer perceptrons for instance), their transformation into more systematic explorations (through the concept of version space, introduced by Mitchel~\cite{mitchell1982generalization} in 1982), the golden age of symbolic learning, and finally, the opposition between supervised and unsupervised learning, so that the classification and clustering problems were at the center of attention in the mid-90s.

Yet it is not described in~\cite{cornuejols2009place}, Probabilistic Graphical Models (PGM) developed in parallel during this period. In fact, this omission is not really surprising, as PGM is a community on its own, well separated from the ML community. In fact, PGM are also classically used in problems such that data fusion or system diagnosis, which may not involve any learning. However, for a long time, Bayesian networks were amongst the state-of-the-art supervised learning methods, along with others non-probabilistic methods ($k$-NN, decision tree, etc.). On the other hand, unsupervised learning was mainly non-probabilistic: AHC, Kohonen maps, and of course, $k$-means.
This latter is interesting: Although not probabilistic, this algorithm provides a relaxed solution of the EM algorithm applied to a mixture of Gaussian models~\cite{dempster1977maximum,kmeansEM}. However, in parallel, Bezdek proposed~\cite{bezdek1984fcm}, and established~\cite{pal1995cluster} a fuzzy version of the $k$-means, named ``fuzzy $c$-means'', which in spite of its non-probabilistic motivation, behaves similarly to the EM algorithm. This clearly illustrates that, in spite of different culture (statistics or fuzzy sets), similar tools are proposed in different communities. Another episode related to the multiple cultural anchors of $k$-means recently showed up: it was established that $k$-means is the discrete counterpart to the PCA~\cite{ding2004k}, a classical non-probabilistic method in statistics, which was developed in the multivariate analysis school~\cite{johnson1992applied}.
In a nutshell, up to the mid-90s, the connections between ML and IA, although weakening, have remained appearent; The problems of interest in ML are related to clustering and classification; Various formalisms are involved in ML, including probabilities, and none of them claims a clear superiority. 

In this context, BFT (still rooted in the probability-affiliated interpretation, while the set-affiliated one is growing) naturally join the trend, so that numerous evidential versions of classical algorithms are successfully proposed and rapidly become state-of-the-art references. Among them, a significant proportion are proposed by Denoeux, such as the evidential $k$-NN~\cite{denoeux1995k} and the evidential neural network~\cite{denoeux2000neural} in the late 90s or the early 2000s.
On this basis, during the early 2000s, numerous derivations of virtually any classical ML algorithms are proposed. Among them, Smets' series on target identification and on Kalman filter~\cite{ristic2004kalman,ristic2004belief,ristic2005target,ristic2005target2,powell2006joint,smets2007kalman} is of prime importance for two reasons. First, they had thrived on top of the generalized Bayesian theorem, proposed a decade earlier~\cite{smets1993belief}. This theorem proposes to link the likelihood and plausibility functions, so that it becomes possible to derive algorithms in the BF framework, which fit the parameters of a model to observed data, in a perfectly compatible ML language~\cite{vannoorenberghe2005partially}. This path was taken over by Denoeux in order to extend Demspter's EM algorithm~\cite{come2008mixture,denoeux2010maximum,denoeux2013maximum}, while cautiously stepping out of the TBM framework.

The second reason of the importance of Smet's series on Kalman filter is cultural: As a defender of his own TBM, he constantly opposed the TBM and the Bayesian views~\cite{ristic2004kalman,smets2005decision}, so that, under his major influence, ML, which from the mid-90s, is less and less influenced by probabilities, was depicted in the BF community mainly as a Bayesian field. In this context, numerous researchers of the BF community, including me (as a PhD student of the mid-2000s), were frozen in a decade-old past, where the ML community would be both inspired by Bayesianism and mainly focused on classification and clustering; two facts that obviously do not hold anymore.

\section{A recap of modern ML}\label{sec:mml}
The works of Vapnik, on support vector machines~\cite{boser1992training} and on statistical learning theory~\cite{vapnik1998statistical}, have provided the foundations of a major revolution which transformed ML in the late 1990s, and which, according to St\'ephane Mallat~\cite{Mallat2014}, still had major impacts fifteen years later. The cornerstones of this revolution are three of them.
The first one is obviously the kernel trick~\cite{scholkopf2001learning}, which deeply roots machine learning in the frameworks of distance geometry, Riemannian analysis, and multivariate analysis (a field of statistics which does not assume any probabilistic model underlying the data). The second one is the idea that a learning problem should be addressed through the minimization of its empirical risk~\cite{erm}. Finally, the third one is to accept that, depending on the dimensionality of the description space, and on the size of the dataset, the empirical risk minimization may be ill-posed, so that a regularizer should be involved in the optimization.

These ideas percolated in the ML community for a decade, providing the tools to give a unifying description~\cite{Mallat2014,coifman2006diffusion,von2007tutorial} of wavelet transform based signal processing, diffusion process, kernel machines and deep neural networks\footnote{The latters have revolutionized computer vision as well as many other ML application fields, as described by De Freitas in a recent keynote lecture~\cite{belief2014}.}. Combined with the first works on variable section by $\ell^1$ penalty~\cite{tibshirani1996regression}, this lead to major breakthroughs in sparse learning~\cite{mairal2010online}, which is of prime importance for the uncertainty theory communities, for its main connection to the \emph{Robust Uncertainty Principles}, proposed by Cand\`es, Romberg and Tao~\cite{candes2006robust}.
 
Todays, in the mid-2010s, ML is not anymore an inter-disciplinary field on which interfere different theories ranging from cognitive sciences through AI to probability, to solve supervised or unsupervised problems: both the background culture and the objectives has changed. Regarding the objectives, they relate to those of information retrieval, social networks, recommender systems, feature extraction, variable selection, data factorization, and sublinear optimization. Even if improving clustering and classification method still deserves some interest\footnote{More precisely, clustering and classification challenging problems still exist, yet, in a setting that differs from the original one on which BFT is classically used: for instance classification of billions of items over millions of classes, or computer vision problems where the classification itself is not the issue with respect to the feature extraction problem that precedes.}, the harder problems presented in ML challenges~\cite{MLchallenges} receives most of the focus.
Regarding the background culture, the field is less interdisciplinary and it is mainly considered as a sub-domain of applied mathematics build on top of optimization (mainly convex), geometry, multivariate statistics and harmonic analysis.
 Very little room is left for subjective probabilities or for AI. This is described in~\cite{cornuejols2009place}, yet it is also well illustrated by the applied mathematics background of most of the researchers recently hired in ML labs.

\section{Some room for belief functions in ML}\label{sec:room}
In this context, it may appear as particularly difficult for the BF community to adapt to the recent evolution of ML, in order to provide state-of-the-art developments. First, BF interpretations and ML have had opposite evolutions so far: From a probability-affiliated view compatible with statistics, BFT had given more room to subjectivism, to finally end in a preponderant set-affiliated view tailored to data fusion and AI problems. On the other hand, ML has quit IA to become more and more tied to functional analysis and convex optimization. This antagonism is well illustrated by the following observation: In ML the entries are raw observations, modeled by a set of points living in a vector space; On the other hand, in the TBM, the entries are assumed to be subjective opinions from different agents and of high semantics. In a similar way, a major asset of the BFT is to provide a rich description of the various types of uncertainty associated with some pieces of information; On the other hand, in ML one is classically not interested in modeling them, but rather to blindly minimize a loss function. Nevertheless, BF community still has several cards to play with respect to ML. In the sequel of this section, I present some of them, sorted according to the BF interpretation they affiliate to.

\subsection{Staying on to set-affiliated interpretations}
The first one is to keep the set-affiliated interpretations, including the TBM, and to  restrict to some very specific ML problems where it is adapted. This is definitely the easiest way, as it prevents any change of the current mainstream of the BF community. However, it remains of interest, even if the path is narrowing under the pressure of big data constraints. Obviously, one must focus on problems where, at the first stage, several agents are used on dedicated learning tasks, and at the second stage, some cooperation or combination between them is expected. This encompasses a wide class of problems that are classically faced in computer vision, among which a few of them~\cite{cuzzolin2013belief,cuzzolin201-pose,reineking2014evidential} are addressed in the BFT:

\begin{itemize}
	\item \textbf{Ensemble learning}: the idea is here to combine the capabilities of several classifiers so that their consensus decision is more robust. This setting has long been explored in the BF framework (see~\cite{kittler1998combining,quost2011classifier} as well as their references), by several researchers, including myself~\cite{burger2006modeling,kessentini2015dempster}. However, the impact all these works is limited by lack of theoretical performance guaranties, such as with boosting methods~\cite{freund1999short}.
	\item \textbf{Co-training}: In this setting~\cite{blum1998combining}, various classifiers work on different feature spaces or on different datasets, in order to have complementary knowledge. Then, each classifier is used to label examples that are useful for the other classifiers to improve their performances. This can be extended to \textbf{transfer learning} problems, where models trained on a first setting are transposed to other similar settings~\cite{courty2014domain}.
	\item \textbf{Active learning}: A classifier asks a human agent to label the training examples that it knows are useful to improve its prediction capability~\cite{settles2010active}.
\end{itemize}
In all these settings, it is required to have several agents, either humans or machines, to automatically evaluate the level of knowledge of each, and to provide a communication scheme between them, so that they can improve one another both their specificity (i.e. being capable of taking a decision), and their consistency (in order to limit their misleading predictions). Described in such a way, BFT seems to be an adapted framework to consider this problem in the most general case. Even if strictly speaking, the learning process would not be accounted for, such a ``cooperation framework'' would be of prime interest for numerous tasks such as complex scene analysis, bioinformatics, etc.

\subsection{Going back to probability-affiliated interpretations}
The second option relies on reversing on purpose the evolution of the BF theories, and to go back to the probability-affiliated interpretations, as they root on solid statistical foundations. Even if ML is more and more involved in optimization, the problem formulation remains in the language of statistics, and as such it is compliant with Dempster's original view\footnote{
Interestingly enough, this couple of short sentences have raised numerous reviewer's comments, among which few of them are worthy being discussed here:
\begin{itemize}
	\item A first comment addressed the fact that in this evolution of ML, the problem formulation remains in the language of statistics, and question its probabailistic interpretation. As a matter of fact, the involvement of a risk function, defined as the expectation of loss function, that is justified to be approximated by the empirical risk on the basis of the assumption that the data are i.i.d., definitely roots machine learning to statistic, multivariate analysis and measure theory; even if traditional hypothesis testing or parametric probabilistic modeling is not involved, and even if optimization methods are used to solve the problem.
	\item Another comment requires being cited: \emph{``Which 'solid foundations'? This assertion seems just the repetition of a clich\'e contrasting the  'solid foundations' of probability and statistics to an alleged lack of such foundations for non-probabilistic approaches. We are never told what these 'solid foundations' actually are.''} The 'solid foundations' of the statistical approaches to ML are described at lenght in Vapnik's \emph{Statistical learning theory}~\cite{vapnik1998statistical} (and that are sketched in the previous bullet), which in addition to providing a axiomatization of the problem, provides metrics to objectively evaluate experimental results as well as generalization bounds. I have never said that non-probabilistic approaches lack of such foundations, as this anonymous reviewer have misread it; yet I believe these foundations have not been proven to adapt to the current ML problems that are already formalized in the literature.
\end{itemize}
}. In such a setting, it could be interesting to address several questions. For instance:
\begin{itemize}
	\item Is it possible to reformulate the  empirical risk minimization principle in the BF framework? Does the minimizer remain convex, if several forms of uncertainties are distinctly accounted for in its expression?
	\item Can we relate the bias-variance dilemma to that of the specificity-consistency trade-off~\cite{pichon2014consistency}? 
	\item Is it possible to extend the multiple test correction setting at the center of most omics studies~\cite{benjamini1995controlling,storey2003statistical} to BF, to account for badly imputed data?
\end{itemize}

Apart from these rather general questions, let us note that the recent works~\cite{come2008mixture,denoeux2010maximum,denoeux2013maximum} which rely on the likelihood interpretation of plausibility, in a setting which differs from the TBM is already a step in this direction. The citation rates of these works acknowledge the idea that going back to a probability-affiliated interpretation makes sense.

\subsection{Moving on to capacity-affiliated or other interpretations}
The last solution is to try to accelerate the evolution of BF community, so that after the probability-affiliated and the set-affiliated interpretations, that of capacity-affiliated focuses the interest. In fact, this view is much more related to optimization, and the options it provides in terms of modeling as well as in terms of solver could definitely be of interest for ML. Basically, it would lead to consider ML learning problems which rely on a game theory setting, or in which multiple criteria have to be optimized in the meantime. Of course, this would naturally lead to models which do not restrict to totally monotone capacities (i.e. BF), but to other types of capacities. More generally, encompassing the BFT in a wider frameworks (Choquet capacities, Walley's lower capacity, or any other, whatever it is) is potentially enriching, and this line has already been adopted by other researchers on questions which are central to ML: Hypothesis testing with interval data~\cite{destercke2014kolmogorov}, regression based on Choquet's capacities~\cite{tehrani2012learning}, partial order ranking~\cite{cheng2010predicting,destercke2013pairwise}, computation of the Vapnik-Chervonenkis dimension of Choquet integral~\cite{hullermeier2012vc}, etc.
Beyond, numerous remaining questions are worthy:
\begin{itemize}
	\item Can we propose an optimizer for the Exploitation-Exploration trade-off (defined in the multi-armed bandit problem~\cite{bubeck2012regret}, as well as in most online learning settings) on the basis of the various imprecision measures available in BFT?
	\item Is there a mean to reduce the imprecision of a source of information, by means of a convex $\ell^1$-penalized optimization, as proposed in~\cite{candes2006robust} for raw signals?
	\item As a penalized optimization amounts to finding a trade-off between two criteria (the loss functions and a regularity measure), does it make sense to  tackle such type of ML problems in the framework of multicriteria optimization?
\end{itemize}

\section{Conclusion}\label{sec:conclusion}
Finally, even if the modern ML mainstream is less interested in the classification/clustering problems that were a natural application field to BFT, there remain few open questions which would advantageously benefit from BFT. Among these open questions, few are based on well-established interpretations which are affiliated to probability or to set theories (cooperation framework), while some requires the BFT to move forwards, to accept wider interpretations (Choquet's capacities) or to be inserted into wider frameworks (lower probabilities). Among these open questions, few are tentatively addressed by leading researchers who points the directions by providing successful first results (imprecise ranking, computer vision, likelihood interpretation of plausibility), while others remain unaddressed (robust uncertainty principle, similarities between bias-variance and specificity-consistency trade-offs). Finally, despite separating from AI to the profit of applied mathematics, ML still provides an interesting playground to BFT researchers. Yet, most importantly, on few specific ML open questions, we can even expect significant BFT contributions.

\bibliographystyle{splncs03}
\bibliography{refs}
\end{document}